\PassOptionsToPackage{unicode}{hyperref}
\PassOptionsToPackage{hyphens}{url}
\PassOptionsToPackage{dvipsnames,svgnames,x11names}{xcolor}
\documentclass[
  11pt,
]{article}
\usepackage{xcolor}
\usepackage[a4paper,margin=25mm]{geometry}
\usepackage{amsmath,amssymb}
\usepackage{iftex}
\ifPDFTeX
  \usepackage[T1]{fontenc}
  \usepackage[utf8]{inputenc}
  \usepackage{textcomp} 
\else 
  \usepackage{unicode-math} 
  \defaultfontfeatures{Scale=MatchLowercase}
  \defaultfontfeatures[\rmfamily]{Ligatures=TeX,Scale=1}
\fi
\usepackage{lmodern}
\ifPDFTeX\else
\fi
\IfFileExists{upquote.sty}{\usepackage{upquote}}{}
\IfFileExists{microtype.sty}{
  \usepackage[]{microtype}
  \UseMicrotypeSet[protrusion]{basicmath} 
}{}
\makeatletter
\@ifundefined{KOMAClassName}{
  \IfFileExists{parskip.sty}{%
    \usepackage{parskip}
  }{
    \setlength{\parindent}{0pt}
    \setlength{\parskip}{6pt plus 2pt minus 1pt}}
}{
  \KOMAoptions{parskip=half}}
\makeatother
\usepackage{color}
\usepackage{fancyvrb}

\DefineVerbatimEnvironment{Highlighting}{Verbatim}{commandchars=\\\{\}}
\newenvironment{Shaded}{}{}

\newcommand{\NormalTok}[1]{#1}

\usepackage{longtable,booktabs,array}
\usepackage{calc} 
\usepackage{etoolbox}
\makeatletter
\patchcmd\longtable{\par}{\if@noskipsec\mbox{}\fi\par}{}{}
\makeatother
\IfFileExists{footnotehyper.sty}{\usepackage{footnotehyper}}{\usepackage{footnote}}
\makesavenoteenv{longtable}
\NewDocumentCommand\citeproctext{}{}
\NewDocumentCommand\citeproc{mm}{%
  \begingroup\def\citeproctext{#2}\cite{#1}\endgroup}
\makeatletter
 \let\@cite@ofmt\@firstofone
 \def\@biblabel#1{}
 \def\@cite#1#2{{#1\if@tempswa , #2\fi}}
\makeatother
\newlength{\cslhangindent}
\newlength{\csllabelwidth}
\newenvironment{CSLReferences}[2] 
 {\begin{list}{}{%
  \setlength{\itemindent}{0pt}
  \setlength{\leftmargin}{0pt}
  \setlength{\parsep}{0pt}
  \ifodd #1
   \setlength{\leftmargin}{\cslhangindent}
   \setlength{\itemindent}{-1\cslhangindent}
  \fi
  \setlength{\itemsep}{#2\baselineskip}}}
 {\end{list}}
\usepackage{calc}

\usepackage{graphicx}
\usepackage{microtype}
\usepackage{xurl}
\usepackage{needspace}
\usepackage{bookmark}
\IfFileExists{xurl.sty}{\usepackage{xurl}}{} 
\hypersetup{
  pdftitle={Agent-Controlled Forgetting for Tool-Using Agents: Reversible Context Curation in Practice},
  pdfauthor={Jan-Peter Franke franke@bold-projekte.com},
  colorlinks=true,
  linkcolor={black},
  filecolor={Maroon},
  citecolor={black},
  urlcolor={blue},
  pdfcreator={LaTeX via pandoc}}

\title{Agent-Controlled Forgetting for Tool-Using Agents: Reversible
Context Curation in Practice}
\author{Jan-Peter Franke\\
\href{mailto:franke@bold-projekte.com}{\nolinkurl{franke@bold-projekte.com}}}
\date{Münster, Germany · 6 October 2026}

\begin{document}
\maketitle
\begin{abstract}
Tool-using agents repeatedly carry observations whose useful content can
be much smaller than their original payload. We study agent-controlled
forgetting: the acting model selects previously observed tool results,
replaces each with a short note at its original position, and retains
the exact original in a recoverable archive. A Python harness exposes
batch archival and explicit recovery without task-specific model
training, while protecting user instructions and assistant messages from
these operations. In an exploratory OpenTelemetry debugging case
followed by an unrelated implementation task, the method ended with
231,951 provider-reported prompt tokens versus 912,492 under retained
history, used 50\% fewer cumulative input tokens, and had an estimated
API cost of \$1.28--\$1.44 versus approximately \$4.38. Both arms passed
the two-case primary behavioral oracle; neither fully satisfied the
follow-up evaluation. The method made more requests and took 17\%
longer. A contrasting application-development pair produced no context
or cost saving, and an earlier continuation exhibited lower manually
assessed quality despite reduced context. These observations demonstrate
substantial resource savings in noisy tool-use trajectories and identify
workload dependence as a central consideration for reversible context
management.
\end{abstract}

\section{Introduction}\label{introduction}

A debugging agent may inspect an entire distributed trace to identify
one failed service call. A coding agent may need the final diagnostic
from a long build log, while the next request still contains every
preceding progress line. These observations can be useful to inspect
once without being useful to retain in full throughout the remaining
task. The resulting research question is not only how much information
an agent can receive, but which already inspected information should
continue to occupy its active context.

The design draws on human practices of organizing working material while
using a software-specific capability: explicitly setting an observation
aside and later restoring its exact contents. An organized desk provides
the analogy: keep current evidence within reach and file completed
material with a short note and a retrieval reference.

We investigate a separation between \textbf{exposure} and
\textbf{continued retention}. After observing a tool result, the same
model that performs the task can decide whether to keep that result
verbatim or replace it with an explicit note. The original remains
recoverable. This decision can use knowledge gained during diagnosis,
including the fact that a previously plausible cause has been ruled out
or that the user has moved to a different task. The mechanism does not
avoid the initial cost of reading a result. Its potential benefit lies
in reducing repeated input and leaving less obsolete material in
subsequent requests.

This distinction matters when the agent cannot change the tools that
produce its observations. A third-party Model Context Protocol (MCP)
server may return a broad response; a browser observation may include
navigation and controls unrelated to the current question; a test runner
may emit extensive logs before a short verdict. Improving those
interfaces or filtering their outputs can reduce initial input. However,
selection before the acting model sees a result may omit unexpected
evidence, and some material becomes dispensable only after the model
interprets it. Post-exposure curation gives the agent a further choice
over continued retention without requiring changes to the source tool.

The mechanism originates in the author's Arios agent, with extraction,
recovery, and explicit forgetting tools recorded in its local
implementation history in February 2026; the companion development
evidence documents this history.

Long-context evaluation has shown that the position of relevant
information can affect retrieval and question-answering performance in
studied models (\citeproc{ref-liu_2024_lost}{Liu et al. 2024}). This
motivates measuring context use, but does not imply a universal token
threshold at which an agent becomes unreliable. A smaller prompt is also
not a measurement of greater reasoning ability. We therefore report
active prompt size, cumulative usage, estimated cost, and task outcomes
separately.

Agent-managed memory is an established research direction. MemGPT
provides model-directed movement between memory tiers; MemAct integrates
context editing into the acting policy; Context-Folding structures
temporary subtrajectories; ACM provides agent-initiated summarization
and archive queries; and AdaCoM trains a separate context manager. Our
contribution is an implementation and exploratory evaluation of a
restricted, auditable operation over individual tool-result occurrences,
not a claim to invent agent-controlled memory. We study how this
operation behaves in real coding and browser trajectories, including a
case in which retained history exceeds 900,000 reported prompt tokens.

The report contributes: (1) a concrete reversible tool-result contract
with stable identifiers, in-place notes, protected instruction messages,
and atomic batch validation; (2) an independently graded
debugging-and-pivot case with provider usage and cache-aware cost
accounting; and (3) contrasting and adverse observations that delimit
where the mechanism is useful. The central case demonstrates substantial
reductions in active context and repeated input while preserving success
on the tested primary behaviors.

\section{Related work}\label{related-work}

\textbf{Memory tiers and agent-directed editing.} Packer et al.
(\citeproc{ref-packer_2023_memgpt}{2023}) introduce MemGPT, where
function calls move information between active and external memory and
recall storage preserves earlier interactions. Its queue manager also
controls overflow. Our evaluated operation targets selected tool outputs
in the current conversation, retaining a note at each original location.
This is a narrower interface choice within the broader area of
agent-managed memory.

\textbf{Learned management policies.} Zhang et al.
(\citeproc{ref-zhang_2026_memory}{2026}) formulate MemAct around a
Prune\&Write operation over identifiable interaction records and train
the policy using Dynamic Context Policy Optimization. Thus, allowing the
task agent to choose context edits is already established. Yi et al.
(\citeproc{ref-yi_2026_learning}{2026}) instead train an external
manager that rewrites, deletes, or merges messages for a frozen acting
agent. Our experiments use a prompted acting model without additional
training or a separate manager. They do not compare against either
learned system.

\textbf{Structured reduction and accumulated knowledge.} Sun et al.
(\citeproc{ref-sun_2025_scaling}{2025}) introduce Context-Folding, which
branches into subtrajectories and returns a summary while removing
intermediate context, with FoldGRPO training. Our operation can select
already completed results without requiring a branch boundary. Zhang et
al. (\citeproc{ref-zhang_2025_agentic}{2025}) describe ACE, where
generation, reflection, and curation update an evolving playbook.
Maintaining reusable knowledge and reducing active observation payload
are complementary goals; we do not evaluate that combination here.

\textbf{Proactive folding and recoverable transcripts.} Ye et al.
(\citeproc{ref-ye_2025_agentfold}{2025}) train AgentFold to combine
folding directives with web-agent actions, maintaining summaries at
multiple scales alongside the latest interaction. This is another
precedent for treating context management as part of agent behavior. The
practical pi-fold implementation (\citeproc{ref-conner_fold}{Conner,
n.d.}) preserves exact transcript spans outside the active window,
replaces them with briefs, and supports expansion. Its inspected README
distinguishes earlier agent-laid marks from a newer automatic folding
policy. Our interface targets individual tool-result occurrences
selected by the acting model and returns recovered payloads as new
results; the shared use of briefs, external originals, and recovery is
not a novelty claim. We have not benchmarked against pi-fold.

\textbf{Retrieval and persistent-memory retention.} Nahid and Rafiei
(\citeproc{ref-nahid_2025_prism}{2025}) introduce PRISM, whose Selector
and Adder iteratively refine an evidence set for multi-hop question
answering. This demonstrates why filtering should not be reduced to a
single irreversible preprocessing step. Wei et al.
(\citeproc{ref-wei_2026_fademem}{2026}) describe FadeMem's adaptive
decay, memory layers, conflict resolution, and fusion. Rusu et al.
(\citeproc{ref-rusu_2026_selective}{2026}) extract conversational
content into a graph and prune nodes using retention scores. These
systems manage retrieval candidates or persistent memory
representations, whereas our measured operation changes the continued
presence of raw observations in an active trajectory.

\textbf{Compression policies and archival.} Kang et al.
(\citeproc{ref-kang_2025_acon}{2025}) introduce ACON, which optimizes
compression guidelines from trajectory feedback and uses a separate
model to compress histories or observations at token thresholds. Our
interface instead lets the acting model retrospectively select
individual tool results and author their replacement notes. Li et al.
(\citeproc{ref-li_2026_acm}{2026}) introduce ACM, which shares
agent-initiated management and archival of raw messages, but summarizes
message intervals with a separate LLM and retrieves query-relevant
extracts through another model. Our contract preserves user, system and
assistant messages and returns exact archived payloads on recovery. ACM
evaluates both an untrained management framework and a post-trained
policy; absence of task-specific training alone is therefore not a
distinguishing feature.

\textbf{Provider-side editing.} Anthropic's context-editing
documentation (\citeproc{ref-anthropic_context_editing}{Anthropic,
n.d.}) describes threshold-triggered removal of older tool results with
placeholders, while the client retains the original conversation. It
also documents cache invalidation when results are cleared. This
provides a practical comparison point for rule-based editing. Our
model-selected notes and explicit recovery interface are choices within
this broader design space; no provider-side baseline is evaluated here.

\textbf{Freely editable context.} Shao et al.
(\citeproc{ref-shao_2026_context}{2026}) introduce Context Language
Models (CLMs), giving the acting model general read-write access to its
live context through a mirrored file, including offloading and
retrieval. Our interface restricts edits to individual tool-result
occurrences, automatically archives exact originals, and preserves user,
system, and assistant messages. Human instructions therefore remain a
verbatim task anchor under our context operations. This preservation
guarantee does not extend to a separate compaction process that
summarizes those messages, nor does it guarantee instruction compliance.
We have not compared against CLM.

\textbf{Positioning.} Pre-exposure retrieval and filtering can reduce
what the agent initially receives. Post-exposure curation can
additionally exploit conclusions reached after inspection. This timing
distinction does not imply that filters are intrinsically inferior:
iterative retrieval, query refinement, and programmatic processing can
also adapt to task state. Our retained-history comparator isolates a
practical policy bundle, not the best available context-management
baseline. The targeted literature review does not establish novelty of
the full combination of archival, notes, identifiers, and recovery.

\section{Method}\label{method}

\subsection{State and eligible
operations}\label{state-and-eligible-operations}

Let the active conversation be an ordered message sequence H, with an
external archive A. An eligible tool-result occurrence contains an
identifier r, tool-call metadata, text x, and optional image payloads v.
Stable identifiers are assigned when results enter the conversation.
They start at r1 in a new session and are never reused within that
session. The result occurrence, rather than its source file or tool
name, is the unit of selection.

The acting model calls \texttt{context\_apply} with a batch of pairs
\texttt{(result\_id,\ note)}. For every selected result, the harness
stores its raw text, images, and provenance in A, then replaces the
active result body with a stub containing the note and archive
reference. The associated tool call remains in the trajectory. A note
may retain an extract, a diagnostic conclusion, source locations, or a
reason why the result is no longer needed.

\texttt{context\_recover(archive\_ref)} returns the original as a new
tool result at the current end of the conversation. The recovered
occurrence receives a new result identifier. Recovery does not rerun the
original tool: an archived browser page or test log remains evidence of
the earlier observation, not a fresh environmental measurement. Original
stubs remain addressable.

The active representation is lossy because a note need not preserve
every detail. Reversibility refers to retrieval of the stored payload,
not to lossless reasoning or preservation of every future decision
dependency. Likewise, forgetting here is eviction from the active
request, not deletion from storage, model unlearning, or a
privacy-erasure mechanism.

\subsection{Integrity and protocol
boundaries}\label{integrity-and-protocol-boundaries}

The backend validates the entire batch before committing a change.
Unknown, duplicate, already archived, or otherwise unavailable
identifiers produce a corrective error without partial mutation.
Originals are archived before stubs are committed. User, system, and
assistant messages are not selectable through the context tools.
Archival does not edit workspace files.

A compact description of the transition is:

\begin{Shaded}
\begin{Highlighting}[]
\NormalTok{apply(batch):}
\NormalTok{    resolve every result ID and validate every note}
\NormalTok{    reject the whole batch if any target is invalid}
\NormalTok{    stage originals and provenance in the archive}
\NormalTok{    commit notes and archive references at the selected result locations}
\NormalTok{    return the result{-}ID to archive{-}reference mapping}

\NormalTok{recover(ref):}
\NormalTok{    load the stored text and image payloads}
\NormalTok{    append them as a new tool result with a fresh result ID}
\end{Highlighting}
\end{Shaded}

The trace preserves original observations independently of the active
representation. This enables comparison of archived payloads against
recorded tool returns. Such equality checks validate storage integrity;
they do not validate the semantic adequacy of notes. The provider
projection also handles API representation constraints, so preservation
of stored assistant responses should not be confused with byte-identical
wire serialization in every error case.

\subsection{Prompted selection}\label{prompted-selection}

The method arm receives context-management instructions, tool
descriptions, examples, and an ephemeral runtime reminder. The acting
model is asked to curate after useful research or diagnostic progress,
batch substantial targets, retain unresolved evidence, and continue the
original task. The prompt mentions roughly 25\% irrelevant content as a
heuristic, together with material size reduction and call overhead. This
percentage is neither an annotated dataset property nor an enforced
classifier threshold.

The final runtime implementation estimates current text size from the
projected request and makes a reminder eligible at approximately 40,000
text tokens, or when image content is present. After a reminder, another
requires meaningful new eligible output, configured as 20,000 characters
or a new image. A new user task resets reminder eligibility. The
reminder is appended as user-role content at the end of the outgoing
request and is not persisted in conversation history. It describes
maintenance rather than a deadline or an instruction to finish.

The condition therefore tests \textbf{prompted, model-selected
curation}. It does not show that an unprompted base model spontaneously
discovers a need to forget, and it does not involve parameter learning
during the experiment. Retained history omits the method instructions,
reminder, and context tools while keeping the shared task tools and
result labels.

\subsection{Resource accounting}\label{resource-accounting}

For request \(t\), let \(P_t\) be provider-reported prompt tokens and
\(O_t\) completion tokens. We report the last measured \(P_t\), maximum
\(P_t\), cumulative input \(\sum_t P_t\), and total reported tokens
\(\sum_t(P_t+O_t)\). The last prompt measurement precedes generation of
the final submission; it is not an exact tokenization of the post-answer
history. Prompt size includes the request representation and is an
operational proxy for active context, not a direct measure of usable
reasoning capacity.

If \(K_t\) cached input tokens are reported, estimated request cost is
\(r_{\mathrm{input}}(P_t-K_t)+r_{\mathrm{cache}}K_t+r_{\mathrm{output}}O_t\).
Missing cache detail produces a range rather than an assumed zero.
Reasoning tokens reported within completion usage are not added again.
For OpenTelemetry, recorded rates were \$0.20 per million ordinary input
tokens, \$0.05 per million cached input tokens, and \$0.50 per million
output tokens. These are run-start tariff estimates, not invoices or
measurements of provider compute.

Curation may invalidate a reusable prompt prefix and increase the cost
of the next request. A smaller subsequent history may nevertheless
reduce total paid input. We measure the resulting usage and tariff
estimate, without assuming either perfect cache preservation or complete
cache loss. Wall time includes tool activity; it cannot identify a
causal time-to-first-token effect.

\section{Experimental design}\label{experimental-design}

\subsection{Scope and development
history}\label{scope-and-development-history}

This is an exploratory systems study using the provider model identifier
\texttt{z-ai/glm-5.3-flash} through TensorX. The principal comparisons
use one fresh session per condition, not a random sample of independent
tasks. Prompt and harness development preceded these comparisons.
Earlier attempts, including unfavorable outcomes and infrastructure
failures, remain recorded. Selecting the high-noise case after earlier
experiments makes it a development-informed case study rather than a
held-out confirmatory benchmark.

We report the repaired OpenTelemetry pair as the central case, the
completed Planroom pair as a workload contrast, and the earlier
six-session Sphinx batch as descriptive supporting evidence. An adverse
browser continuation is included explicitly. The experiments are not
pooled into an overall accuracy score: tasks, graders, histories, and
harness versions differ.

\subsection{OpenTelemetry debugging and task
pivot}\label{opentelemetry-debugging-and-task-pivot}

The environment uses the public OpenTelemetry Demo source before an
upstream checkout fix, at commit
\texttt{af3dc6553d9f7aebb5929b192fbf3a922a467050}. The primary task is
to diagnose a failing checkout using browser and observability surfaces,
implement a safe frontend error contract, and verify it. Background load
and concurrent payment, advertising, image, Kafka, homepage, and
product-catalog faults produce realistic competing observations. No
prose padding or mandatory reading volume is injected. The fraction of
irrelevant material is not independently annotated.

After primary submission, the same session receives an unrelated C++
currency-service task requiring graceful shutdown. Each condition
carries its own actual primary history and workspace into this phase.
Consequently, the follow-up compares full trajectories, not an
identical-history intervention that isolates the effect of the pivot
itself.

Retained runs first and method second. Each starts from a reset stack
and a byte-identical source snapshot. Shared model configuration, task
contracts, application setup, and task tools are held fixed; curation
instructions and tools differ by design. Live telemetry, browsing
choices, and generated code are not identical across conditions. The
runner configures 1,500 logical calls and 1,800 provider attempts per
episode, with a 131,072-token output budget and requested low reasoning
effort. Neither condition reaches those limits. No fixed provider
sampling seed or explicit temperature is supplied in this runner.

The private primary oracle rebuilds a copy of the candidate and runs two
upstream-derived Cypress cases. Both baseline failure and gold success
were checked. Passing those cases supports the tested error behavior,
not exhaustive satisfaction of every prose requirement.

The currency grader combines compilation, process probes, and
source-text checks. It tests SIGTERM and SIGINT, process identity, exit
status, elapsed time, and an occupied-port startup failure. Its
six-criterion summary additionally uses string and regular-expression
checks for implementation structure and reporting. The baseline scores
0/6 and the reference 6/6. However, calibration on one reference cannot
establish general grader validity: alternative correct implementations
may fail static patterns, and matching a pattern does not prove
semantics. We therefore emphasize executable observations and treat the
aggregate follow-up score as provisional supporting evidence.

Thirty-seven benchmark inputs and a 605-file snapshot were frozen before
the final pair. Version 01 had a compromised follow-up build surface and
an incompletely specified primary error contract. A retained-only
version-02 development episode exposed further deadline-contract
ambiguity. Those trajectories are retained but excluded from the
version-03 comparison. No post-grading repairs or selective reruns were
applied to its candidates.

\subsection{Contrasting tasks}\label{contrasting-tasks}

\textbf{Planroom.} A multi-phase application-development task requires
library-documentation research through Context7, implementation, and
twelve prescribed browser journeys through Playwright. Both arms use the
same specification and scaffold, with a completion gate binding
criterion identifiers to their full requirement text. Journey coverage
is assessed from traces and independent checks. Model-written test
counts differ and are not comparable accuracy denominators. File-picker
behavior and deliberate persistence-failure injection remain evaluation
gaps.

\textbf{Sphinx publication manifest.} Six fresh sessions receive three
successive implementation requests on Sphinx 8.2.3. The predefined order
is retained, method, method, retained, retained, method. Private graders
contain 11, 20, and 24 cases across phases. This older experiment uses a
16,384-token output limit and a global 300-request guard. Its different
configuration prevents treatment as a replication of the final
OpenTelemetry interface. Cases within each grader share requirements and
are not independent statistical samples.

\Needspace{24\baselineskip}

\section{Results}\label{results}

\begin{center}\includegraphics[width=\linewidth]{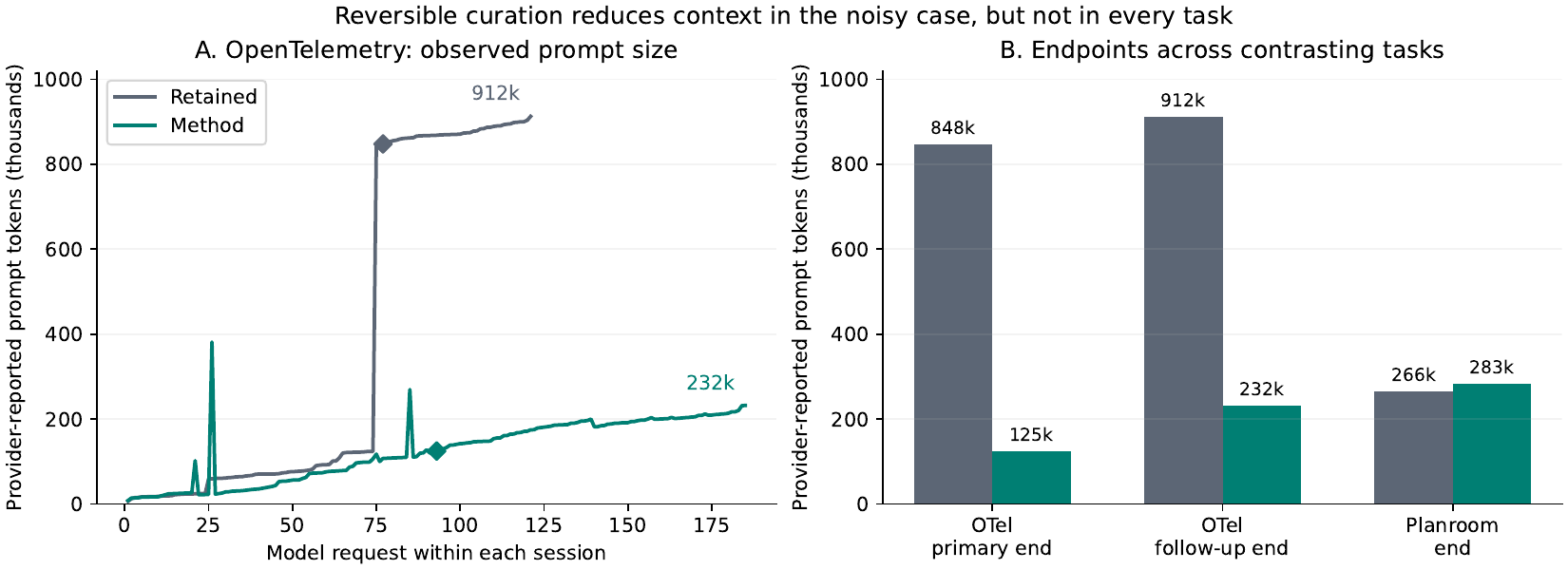}\end{center}

\emph{Figure 1. Provider-reported prompt size. Panel A plots each
OpenTelemetry session against its own request index; diamonds mark
primary submission, so equal horizontal positions do not denote equal
work. Panel B includes the contrasting Planroom result. These are
individual trajectories, with no uncertainty bands. The plotted endpoint
is the input to the final response, not a post-answer context
measurement.}

\subsection{OpenTelemetry resource
use}\label{opentelemetry-resource-use}

Both arms submit both tasks normally, without recorded provider errors.
Table 1 shows substantial reduction in active and cumulative input under
the method despite more requests. At primary submission the method
carries 124,746 prompt tokens versus 847,674 for retained, an 85\%
reduction. After the follow-up, the method carries 231,951 versus
912,492 prompt tokens, a 75\% reduction. The method peak is 381,299
versus 912,492 tokens. Thus, the observed comparison reaches a
substantially larger retained context than the earlier coding tasks.

\emph{Table 1. OpenTelemetry, both phases unless specified.}

{\def\LTcaptype{none} 
\begin{longtable}[]{@{}
  >{\raggedright\arraybackslash}p{(\linewidth - 4\tabcolsep) * \real{0.5435}}
  >{\raggedleft\arraybackslash}p{(\linewidth - 4\tabcolsep) * \real{0.2283}}
  >{\raggedleft\arraybackslash}p{(\linewidth - 4\tabcolsep) * \real{0.2283}}@{}}
\toprule\noalign{}
\begin{minipage}[b]{\linewidth}\raggedright
Metric
\end{minipage} & \begin{minipage}[b]{\linewidth}\raggedleft
Retained
\end{minipage} & \begin{minipage}[b]{\linewidth}\raggedleft
Method
\end{minipage} \\
\midrule\noalign{}
\endhead
\bottomrule\noalign{}
\endlastfoot
Primary behavioral oracle & 2/2 & 2/2 \\
Primary submission prompt tokens & 847,674 & 124,746 \\
Final measured prompt tokens & 912,492 & 231,951 \\
Peak prompt tokens & 912,492 & 381,299 \\
Cumulative input tokens & 45,766,146 & 22,988,240 \\
Completion tokens & 33,834 & 71,919 \\
Input plus completion tokens & 45,799,980 & 23,060,159 \\
Model requests & 121 & 185 \\
Elapsed seconds & 1,730 & 2,025 \\
Estimated API cost, USD & 4.378--4.379 & 1.283--1.436 \\
Successful archival batches & 0 & 9 \\
\end{longtable}
}

Cumulative input decreases by 50\%, and input plus completion tokens
decrease by 50\%. The estimated cost is approximately 67--71\% lower,
even after reported cache usage is included. The method generates more
than twice as many completion tokens and takes 17\% longer. These are
resource savings without a speed advantage in this case. Original
ledgers and exact deltas are linked in the OpenTelemetry metrics.

The largest retained-context jump follows a 2,478,818-character
\texttt{browser\_find} result at primary step 74, three requests before
primary submission; the follow-up then carries the accumulated history
through 44 further requests. Because the arms encounter different
observations and most primary implementation precedes this jump, the
pair measures complete retention trajectories rather than controlled
exposure to identical noise.

\subsection{Observed task quality}\label{observed-task-quality}

Both primary candidates pass the two-case behavioral oracle.

For the follow-up, both candidates compile. The method exits with status
0 after SIGTERM in 8.43 seconds and after SIGINT in 8.85 seconds.
Retained fails both probes and is killed with status 137. Conversely,
retained passes the occupied-port startup-failure probe with status 1;
the method's second process does not finish within 30 seconds. The
method therefore shows stronger tested signal handling but weaker tested
startup-failure handling.

The frozen mixed grader reports 3/6 for method and 2/6 for retained.
Source inspection identifies false negatives in its static checks:
method calls \texttt{server\_ptr-\textgreater{}Shutdown(...)} rather
than the expected \texttt{server-\textgreater{}Shutdown(...)}, and
retained uses \texttt{signalfd} rather than the expected
\texttt{sigwait}. Both candidates contain remaining-budget calculations.
Method's startup-failure branch joins a signal-waiting thread without
first arranging its termination, consistent with the observed timeout.
The frozen grades are preserved, but their mixture of process probes and
syntax-sensitive checks makes the individual runtime outcomes more
informative than the aggregate score. See the retained and method
records.

\subsection{What was archived}\label{what-was-archived}

The method makes five successful archival batches in the primary phase
and four in the follow-up. In total, 26 results containing 2,247,229
text characters are represented by 9,709 note characters. Archived
material includes large observability responses and completed diagnostic
or verification output. All 26 archived text payloads and image lists
match their recorded tool returns exactly. No recovery is used in this
pair, and no image payload is archived.

The archive verifies exact payload preservation while the agent
repeatedly curates context and continues working; this pair does not
test retrieval behavior because no recovery is requested.

\Needspace{16\baselineskip}

\subsection{Workload contrast and repeated
pilot}\label{workload-contrast-and-repeated-pilot}

\emph{Table 2. Planroom completed pair.}

{\def\LTcaptype{none} 
\begin{longtable}[]{@{}lrr@{}}
\toprule\noalign{}
Metric & Retained & Method \\
\midrule\noalign{}
\endhead
\bottomrule\noalign{}
\endlastfoot
Supported browser journeys & 12/12 & 12/12 \\
Final and peak prompt tokens & 265,965 & 283,105 \\
Input plus completion tokens & 67,669,624 & 68,136,934 \\
Model requests & 442 & 427 \\
Elapsed minutes & 31.78 & 24.29 \\
Estimated API cost, USD & 3.461--3.465 & 3.510--3.577 \\
\end{longtable}
}

Planroom yields no context or cost saving: the method ends with 6\% more
prompt tokens (283,105 versus 265,965) and uses less than 1\% more total
tokens (68,136,934 versus 67,669,624). Both arms support the twelve
journeys at the available assessment resolution, with the gaps described
above. Archival of 34 results removes only 69,685 net history
characters; this does not offset the differing generated trajectories
and method overhead. The method completes in 24.29 rather than 31.78
minutes; this single pair does not isolate the cause of the timing
difference. Planroom evidence.

\emph{Table 3. Sphinx pilot, all six sessions in execution order.}

{\def\LTcaptype{none} 
\begin{longtable}[]{@{}
  >{\raggedright\arraybackslash}p{(\linewidth - 10\tabcolsep) * \real{0.1364}}
  >{\raggedright\arraybackslash}p{(\linewidth - 10\tabcolsep) * \real{0.1364}}
  >{\raggedleft\arraybackslash}p{(\linewidth - 10\tabcolsep) * \real{0.1818}}
  >{\raggedleft\arraybackslash}p{(\linewidth - 10\tabcolsep) * \real{0.1818}}
  >{\raggedleft\arraybackslash}p{(\linewidth - 10\tabcolsep) * \real{0.1818}}
  >{\raggedleft\arraybackslash}p{(\linewidth - 10\tabcolsep) * \real{0.1818}}@{}}
\toprule\noalign{}
\begin{minipage}[b]{\linewidth}\raggedright
Session
\end{minipage} & \begin{minipage}[b]{\linewidth}\raggedright
Condition
\end{minipage} & \begin{minipage}[b]{\linewidth}\raggedleft
Final private cases
\end{minipage} & \begin{minipage}[b]{\linewidth}\raggedleft
Final prompt
\end{minipage} & \begin{minipage}[b]{\linewidth}\raggedleft
Total tokens
\end{minipage} & \begin{minipage}[b]{\linewidth}\raggedleft
Requests
\end{minipage} \\
\midrule\noalign{}
\endhead
\bottomrule\noalign{}
\endlastfoot
01 & Retained & 23/24 & 267,349 & 15,946,799 & 116 \\
02 & Method & 24/24 & 89,855 & 8,286,503 & 152 \\
03 & Method & 24/24 & 145,213 & 16,849,371 & 209 \\
04 & Retained & 24/24 & 268,937 & 20,217,639 & 136 \\
05 & Retained & 23/24 & 234,057 & 12,273,223 & 98 \\
06 & Method & 23/24 & 123,258 & 12,995,595 & 192 \\
\end{longtable}
}

The Sphinx batch contains two final 24/24 method outcomes and one
retained 24/24 outcome. Every method session has a smaller final prompt
than every retained session, while cumulative usage overlaps. These six
sessions characterize variation on one authored task under the older
configuration, rather than general non-inferiority. Sphinx metrics.

\subsection{Adverse and compromised
observations}\label{adverse-and-compromised-observations}

An earlier browser-based workflow-design continuation used 63\% fewer
total tokens with the method but received a manual score of 4/12 versus
6/12 for retained. Both answers had substantive state-machine errors. An
explicitly requested method-only repetition later scored 6/12. That
repeat was selected after observing the weak answer and must not replace
it or count as a balanced replication. The original result therefore
remains an observed quality loss alongside a resource saving. Adverse
result, selected repeat.

Earlier development also encountered step caps, truncated output,
invalid historical tool arguments, provider failures, and a faulty
benchmark build surface. These failures explain subsequent harness
changes and are not evidence of model degradation caused by context
size. They remain part of the experiment history and are summarized in
the companion evidence map.

\section{Discussion}\label{discussion}

\textbf{Selection after diagnosis.} Large tool outputs can be necessary
for discovering an explanation but redundant after that explanation is
established. The OpenTelemetry archive illustrates that the agent can
act on this distinction. Repeated input savings arise only after the
initial exposure. If subsequent work is short, or if most observations
remain relevant, the overhead of extra prompts, notes, and tool calls
can outweigh the reduction. Planroom provides a concrete boundary case.

\textbf{Working with verbose external tools.} The retention operation
acts on recorded observations, so an upstream tool need not expose its
own compaction controls. In principle, the same contract can cover
documentation, diagnostic logs, browser text, and images. Its value
depends on whether the agent can retain the useful evidence after
inspection, not merely on the size of a response. Filtering and curation
can be combined: one limits initial exposure, while the other revisits
retention as the task evolves.

\textbf{Task pivots.} A new user request supplies information about what
should remain active. For example, after a diagnosis is complete, the
agent can retain the finding and its source references while archiving
the long observations used to establish it. Batch selection permits a
separate note for each result, preserving the surrounding conversation
and access to the originals rather than replacing the entire session
with a single summary. The same-session continuation exercises the
mechanism beyond one isolated answer. It shows that the method can
continue implementing after context edits while retained history
continues to grow. An identical-history branching comparison is the next
step for isolating the effect of curation at the pivot itself.

\textbf{Why retained history first, and context editing next.} Retained
history answers the initial deployment question: what changes when an
acting agent gains explicit control over which tool results remain
active? It measures the combined interface and prompting policy against
keeping all observations. The next direct comparison should test this
policy against threshold-triggered removal of older tool results, as
documented in Anthropic's context-editing documentation
(\citeproc{ref-anthropic_context_editing}{Anthropic, n.d.}). That
comparison would ask whether task-aware selection, per-result notes, and
recovery improve the resource--quality trade-off beyond automatic
clearing. A controlled test should use the same acting model, tools,
tasks, and output budget across conditions; a harness implementation of
an analogous clearing policy must be distinguished from a test of the
provider's actual feature. This baseline is a priority for the next
study and has not been evaluated in the present experiments.

\textbf{Cache and control.} The cost result is compatible with reduced
input outweighing cache disruption, but it does not identify the
provider's cache mechanics. More model decisions can coexist with lower
total cost, as in OpenTelemetry, or with a small cost increase, as in
Planroom. Curation frequency is consequently not a useful success metric
on its own. ACON includes acting-agent and compressor calls in its
practical API cost estimate but explicitly excludes input caching
(\citeproc{ref-kang_2025_acon}{Kang et al. 2025}). Our accounting
includes reported cache usage; the different workloads and models
preclude a direct numerical comparison.

\textbf{Protected instructions and remaining risks.} Protecting user
messages prevents direct archival of the task text. It does not ensure
that the agent follows those instructions or preserves the evidence
needed to satisfy them. Notes can be incomplete, mistaken, or overly
confident; the model may fail to notice when recovery is necessary.
Exact archive integrity addresses only one part of that problem.

\section{Limitations and next
experiments}\label{limitations-and-next-experiments}

The principal limitation is sampling: one OpenTelemetry pair and one
Planroom pair do not estimate the variance of the policy's effect. The
Sphinx batch has three sessions per arm but only one authored task and
an older configuration. There is no preregistered held-out task
distribution, no randomized OpenTelemetry order, and no evidence for a
universal degradation threshold. Confidence intervals over requests or
individual grader cases would incorrectly treat dependent observations
as independent replicates.

The method condition includes instructions, examples, reminders, and
tools. No ablation separates these components, and no comparison tests
provider-side context editing, a learned manager, chronological
eviction, summarization, or programmatic filtering. The experiments
therefore establish neither superiority over other context-management
systems nor novelty of the shared archival and recovery design. Task
trajectories also differ: retained OpenTelemetry has 275 logical image
occurrences in requests while method has none. These are repeated input
occurrences, not 275 unique screenshots. Provider usage is aggregate and
does not expose a separate image-cost decomposition. We cannot attribute
the entire between-arm resource difference to text archival alone.

The archive audit establishes byte-exact preservation, not the
sufficiency of every note or the dispensability of every removed result.
With no recovery in the main pair, its effect on model decisions and the
reliability of downstream retrieval remain unmeasured. The browser
continuation also has different inherited histories and unblinded coarse
grading. Smaller active context alone is not a measure of greater
reasoning ability or a guarantee of preserved quality.

Quality measurement is uneven. The primary oracle covers two specific
behaviors; the currency score combines dynamic checks with
syntax-sensitive static heuristics; Planroom includes trace-based
assessment; and the browser continuation is manually scored. Shared
checks and requirements make individual grader cases dependent; passing
the two-case primary oracle does not establish complete task correctness
or statistical equivalence. Remaining-budget code in both follow-up
solutions does not by itself establish a correct end-to-end timeout
bound. Gold/baseline separation validates the instrument against its
reference candidates but does not eliminate these limits. Before
interpreting follow-up score differences as capability improvements,
broader implementation-agnostic behavioral tests are needed.

The system is evaluated through one provider model identifier.
Provider-side checkpoint identity, inference hardware, cache internals,
energy use, and training-data overlap with public repositories are
unavailable. Run-start configuration and tariffs are recorded, but exact
model and environment determinism is not guaranteed. The hosted
inference figures exclude local build, browser, evaluation, and
researcher labor costs.

The next study should freeze the corrected setup, run balanced repeated
sessions with randomized order, and define task-level outcomes before
inspection. An identical-history pivot experiment would isolate
retention changes from earlier trajectory differences. A recovery-demand
task should make previously archived details necessary and measure
retrieval success and downstream correctness. Finally, direct
comparisons against compacting and filtering baselines, including ACON
and ACM, are required to evaluate the practical value of the specific
interface, rather than the value of context reduction in general.

\section{Conclusion}\label{conclusion}

Agent-controlled forgetting provides an explicit way for a tool-using
model to distinguish information it needed to inspect from information
it still needs to carry. In the central noisy debugging case, reversible
curation accompanies a 75\% smaller final measured prompt, 50\% less
cumulative input, and substantially lower estimated API cost, while both
arms pass the same two primary behavioral cases. The method also
requires more actions and time, and follow-up correctness remains
incomplete. Contrasting application development and an adverse
continuation show that neither resource savings nor quality preservation
follows automatically from a smaller active history. The practical
result is that an agent can complete noisy tool-driven work with
substantially less active context and repeated input, using explicit
archival decisions during the task. The contrasting outcomes define
where further comparisons and replication are needed.

\section{Reproducibility, provenance, and
disclosure}\label{reproducibility-provenance-and-disclosure}

The local workspace preserves prompts, task snapshots, request traces,
archives, submitted workspaces, tariff ledgers, and independent grading
artifacts. The OpenTelemetry manifest records the frozen benchmark and
input hashes. The evidence map specifies the artifact behind each
central claim. The author is Jan-Peter Franke, based in Münster,
Germany. The companion research artifacts are available in the
\href{https://github.com/boldprojekte/agent-forgetting-research}{public
repository}. It includes the main-case harness and benchmark inputs,
submitted candidates, and numerical evidence for the reported
comparisons. Full provider transcripts and third-party literature copies
are not part of that release package. Author-owned research code is
released under the MIT license and derived numerical records under CC BY
4.0; third-party code retains its original licenses. The paper and its
original figures are licensed under
\href{https://creativecommons.org/licenses/by/4.0/}{CC BY 4.0}.

The harness and this manuscript were developed with AI assistance under
the author's direction. AI assistance was also used in experiment
construction and analysis, so author review of interpretation and
evaluation code is required before submission. No model training was
performed for this study. Archival preserves information rather than
erasing it, and a deployed system must account for that persistence when
handling sensitive inputs. We do not infer energy savings from token or
price reductions.

\small
\interlinepenalty=10000

\section*{References}\label{references}
\addcontentsline{toc}{section}{References}

\protect\phantomsection\label{refs}

\end{document}